\documentclass{article}

\usepackage{arxiv}

\usepackage[utf8]{inputenc} 
\usepackage[T1]{fontenc}    
\usepackage{hyperref}       
\usepackage{url}            
\usepackage{booktabs}       
\usepackage{amsfonts}       
\usepackage{amsmath}        
\usepackage{nicefrac}       
\usepackage{microtype}      
\usepackage{lipsum}		
\usepackage{graphicx}
\usepackage{natbib}
\usepackage{doi}

\title{Comparative Analysis on Snowmelt-Driven Streamflow Forecasting Using Machine Learning Techniques}

\author{ \href{}
    {\hspace{1mm}UKESH THAPA}\\
	AI Research Center\\
	Advanced College of Engineering and Management,\\ 
        Tribhuvan University\\
	Kathmandu, Nepal\\
	\texttt{ukesh.thapa@acem.edu.np}\\
	\And
        {\hspace{1mm}Bipun Man Pati} \\
	AI Research Center\\
	Advanced College of Engineering and Management,\\ 
        Tribhuvan University\\
	Kathmandu, Nepal\\
	\texttt{bipunmanpati@acem.edu.np} \\
	\And
        {\hspace{1mm}Samit Thapa} \\
	Department of Civil Engineering\\
	Advanced College of Engineering and Management,\\ 
        Tribhuvan University\\
	Kathmandu, Nepal\\
	\texttt{samit.thapa@acem.edu.np} \\
	\And
        {\hspace{1mm}Dhiraj Pyakurel} \\
	Department of Electronics and Computer Engineering\\
	Advanced College of Engineering and Management,\\ 
        Tribhuvan University\\
	Kathmandu, Nepal\\
	\texttt{dhiraj@acem.edu.np} \\
	\And
        {\hspace{1mm}Anup Shrestha} \\
	Department of Electronics and Computer Engineering\\
	National College of Engineering,\\ 
        Tribhuvan University\\
	Lalitpur, Nepal\\
	\texttt{anup@nce.edu.np} \\
}

\renewcommand{\shorttitle}{Comparative Snowmelt-Driven Streamflow Forecasting}

\hypersetup{
pdftitle={Comparative Analysis on Snowmelt-Driven Streamflow Forecasting Using Machine Learning Techniques},
pdfauthor={Ukesh Thapa, Bipun Man Pati, Samit Thapa, Dhiraj Pyakurel, Anup Shrestha},
pdfkeywords={Support Vector Regression (SVR), Long Short Term Memory (LSTM), Transformer, Temporal Convolutional Network (TCN), Nested Cross-Validation, Snowmelt},
}

\begin{document}
\maketitle

\begin{abstract}
The rapid advancement of machine learning techniques has led to their widespread application in various domains including water resources. However, snowmelt modeling remains an area that has not been extensively explored. In this study, we propose a state-of-the-art (SOTA) deep learning sequential model, leveraging the Temporal Convolutional Network (TCN), for snowmelt-driven discharge modeling in the Himalayan basin of the Hindu Kush Himalayan Region. To evaluate the performance of our proposed model, we conducted a comparative analysis with other popular models including Support Vector Regression (SVR), Long Short Term Memory (LSTM), and Transformer. Furthermore, Nested cross-validation (CV) is used with five outer folds and three inner folds, and hyper-parameter tuning is performed on the inner folds. To evaluate the performance of the model mean absolute error (MAE), root mean square error (RMSE), R square ($R^{2}$), Kling-Gupta Efficiency (KGE), and Nash-Sutcliffe Efficiency (NSE) are computed for each outer fold. The average metrics revealed that TCN outperformed the other models, with an average MAE of 0.011, RMSE of 0.023, $R^{2}$ of 0.991, KGE of 0.992, and NSE of 0.991. The findings of this study demonstrate the effectiveness of the deep learning model as compared to traditional machine learning approaches for snowmelt-driven streamflow forecasting. Moreover, the superior performance of TCN highlights its potential as a promising deep learning model for similar hydrological applications.

\end{abstract}


\section{Introduction}
\subsection{Hindu Kush Himalayan Region}

Snowmelt is the process by which the snow or ice on the ground or various surfaces changes into water due to the temperature rise. Snowmelt plays a pivotal role in the environment and human society development. Snowmelt is also a major source of fresh water in the world. Central Asia contains a large amount of ice in the Hindu Kush-Himalayan (HKH) region and is highly sensitive to global climate change, facing significant warming (0.21±0.08$^{\circ}$/ decade) over the past few decades. In \cite{panday2011detection} a dynamic threshold-based method was applied and observation was done from (2000-2008), where average melt duration was calculated and longer melt season ($\sim$ 5 weeks) occurred in the eastern Himalayan region relative to central, western Himalayan, and the Karakoram region. People residing in the river basins originating from the snow-dominated HKH region rely on rivers for food, water supply, and electricity (\cite{wester2019hindu}). The seasonal snowmelt process significantly affects the ecosystem, water availability, agriculture, and water resources. Additionally, the snowmelt directly influences the river flow, aquatic habitats, and the functioning of the hydroelectric power system. Therefore, precise forecasting of the snowmelt runoff is essential in this area for efficient planning and management. The geographical conditions and extreme weather in this region lead to inadequate ground-truth information for developing the optimum water resource utilization strategy. Therefore, remotely sensed snow and meteorological data are indispensable tools for traditional ground-based monitoring.

\subsection{Snowmelt Forecasting Using Machine Learning}

Machine learning (ML) has diverse applications across interdisciplinary fields, showcasing its versatility and impact. The application of ML in the domain of water resources has been well investigated in previous studies (\cite{asce2000artificial, asce2000artificialII}). \cite{callegari2015seasonal} proposes Support Vector Regression (SVR) for snowmelt-driven discharge forecasting in the Italian Alps using Snow Cover Area (SCA), antecedent discharge (Q), and metrological data, which has outperformed a simple linear auto-regressive model with an average of 33\% relative root mean square error (RMSE) on test samples. \cite{de2018operational} demonstrates the applicability of the SVR model in operational river discharge forecasting, where the SVR model performed better on a single gauging station with a mean improvement of about 48\% in the RMSE. In a study, \cite{uysal2016improving} used a simple Artificial Neural Network (ANN) for snowmelt runoff prediction, where SCA, Q, temperature (T), and precipitation (P) were used as inputs to the model, and research verified that the ANN model outperformed any temperature index (TI) model as its Nash-Sutcliffe model efficiency (NSE) increased from 0.51 to 0.71 in forecasting. In a study by \cite{najafzadeh2023ecological}, an improved version of the robust ML models is proposed to compare the performance of the gene-expression programming (GEP) (R = 0.6089 and RMSE = 7.7229), evolutionary polynomial regression (EPR) (R = 0.6020 and RMSE = 1.6923), multivariate adaptive regression spline (MARS) (R = 0.4333 and RMSE = 11.5046), and model tree (MT) to find the ecological status of the river. In \cite{najafzadeh2023long} has used three different ML techniques such as GEP, MT, and MARS for streamflow forecasting; they claimed that MT had lower uncertainty (95\%PPU = 0.59 and R-factor = 1.67) than other models. 

In past research studies, traditional ANNs were used for the time series problem, but it was not enough to capture all the necessary or complex information from the data. The advancement of technology leads to the development of more complicated architectures. At first, a study proposed by \cite{nagesh2004river} used a Recurrent Neural Network (RNN) for the time series problem, but due to its drawbacks, like the exploding and vanishing gradient problems, Long Short-Term Memory (LSTM) overcame all of those problems. In their research, \cite{kratzert2018rainfall} used two-layered LSTM for rainfall-runoff modeling with NSE 0.63 and argued that it can also mimic the snowmelt process by learning the relationship between precipitation during winter and runoff in spring, but they did not investigate the influence of the hyper-parameter tuning on the model performance. \cite{le2019application} proposes an LSTM model that has an average error of 150 $m^{3}/s$ RMSE with NSE above 99\% while forecasting discharge for one day. \cite{le2019application} and \cite{fan2020comparison} have claimed that window size in forecasting problems is an important hyper-parameter to be tuned for the best model performance, but in both studies, other hyper-parameters, such as the number of LSTM layers and optimizers, were not observed. In \cite{granata2022stacked}, have done a comparative analysis of random forest (RF), SVR, and Multi-layered perceptron (MLP) for the estimation of daily volumetric soil water content, where MLP has the best performance with Coefficient of determination ($R^{2}$) value 0.957. Similarly, In \cite{najafzadeh2024vulnerability} a comparative analysis of nine different ML models is done to predict the vulnerability of the rip current event using two parameters such as dimensionless fall velocity parameter ($\Omega$) and tide range (TR). In a recent study, \cite{thapa2020snowmelt} has compared snowmelt prediction on different models by hyper-tuning parameters with the hit and trail method and setting window size for prediction. Also, gamma testing was done where SCA, T, and Q were chosen as inputs for different models, out of which the LSTM model has better performance with 0.997, 0.112, 0.173, 0.99, and 0.995 as $R^{2}$, mean absolute error (MAE), RMSE, Kling-Gupta efficiency (KGE), and NSE respectively. This study did not evaluate the performance of the model using nested cross-validation (CV), the state-of-the-art (SOTA) model created only performed better for the specified dataset but not for any unseen dataset.

The main objective of this study is to investigate the SOTA deep learning (DL) architectures such as the Transformer and the Temporal Convolutional Networks (TCN) to perform snowmelt prediction over traditional ML i.e. SVR and old DL techniques i.e. LSTM. In the snowmelt prediction, the relationship between the input and the target variable can be complex and dynamic, so we compare the performance of the traditional ML and DL techniques to handle the complex pattern of the data for forecasting. In the current literature reviews, we found no consistent metrics are being used to compare model improvement. We have compared the four different models and have used consistent metrics (such as KGE, NSE, RMSE, and MAE) to create a benchmark for the model to compare its performance. Furthermore, we evaluate the performance of each model using nested CV for the model generability test. In this study, SCA, T, Q, and remotely sensed daily P are used as inputs. The remainder of the paper is organized as follows: Section 2 provides the materials and methods of the study. Section 3 then describes the methodology. The results of the proposed study are then discussed in Section 4. The discussion part is discussed in Section 5. Finally, Section 6 provides the conclusion and directions for future research.

\section{Materials and Methods}
\subsection{Study Area}
\begin{figure}[h]
    \begin{center}
    \includegraphics[width=6.5cm]{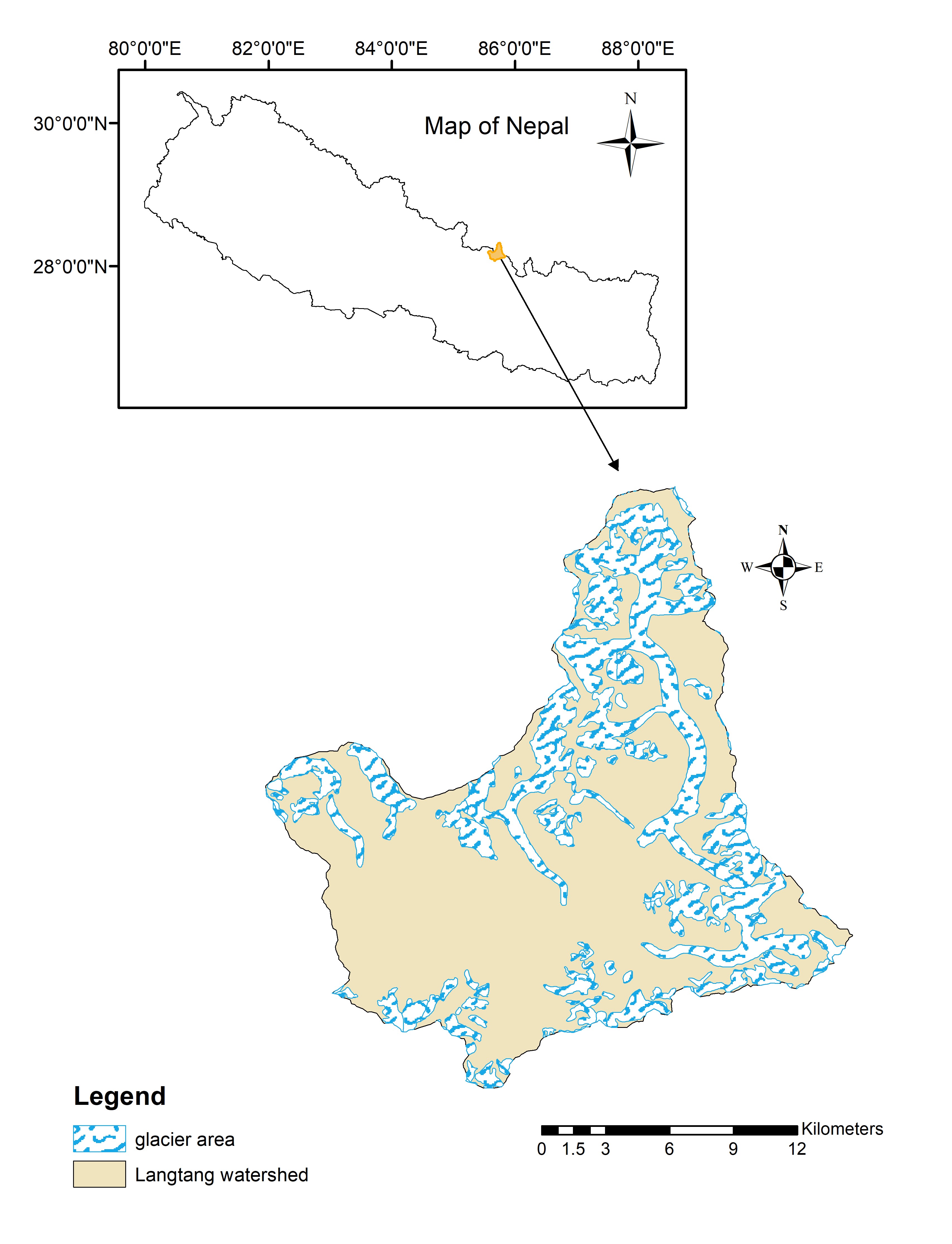}
    \end{center}
    \caption{ Map of Langtang basin area}
    \label{fig:1}
\end{figure}

In the Langtang basin, which is situated in Nepal's Central Himalayas, the study's main focus is on predicting the runoff from snowfall. Glaciers occupy 110 km² of the 354 km² study area, which spans from 3647 to 7213 meters above sea level. RGI-GLIMS version 6 calculates the extent of glaciers (\cite{rgi2017dataset}). The entire amount of water discharge in this watershed is largely determined by the melting of glaciers and snow (\cite{ragettli2015unraveling}). Figure \ref{fig:1} shows the HKH area which is used for the dataset collection.
\subsection{Hydrometeorological Data}
In Nepal, the Department of Hydrology and Meteorology handed the hydrological data from the Kyangjing station ($28.216^\circ$ latitude, $85.55^\circ$ longitude). The APHRODITE product (APHRO\_TAVE\_MA\_V1808) handed diurnal temperature data with a $0.25^\circ \times 0.25^\circ$ grid for the times 2002--2012 [\cite{yasutomi2011development}]. Previous exploration in the Langtang [\cite{thapa2020trend}] receptacle revealed a high correlation (Spearman's $\rho > 0.9$) between APHRODITE products and ground compliances. The rush data were attained using 3B42RT TRMM lines, which were sourced from the Tropical Rainfall Measuring Mission (TRMM) at a resolution of $0.25$ degrees. These datasets combine information from multiple detectors including satellite infrared data to estimate rush. The 3B42RT TRMM dataset is created by combining data from rain needles. TRMM products are available at \href{https://pmm.nasa.gov}{https://pmm.nasa.gov}  and are constantly used in the Himalayan region (\cite{immerzeel2009large}).

\subsection{Snow Cover}
The MODIS MOD10A2 interpretation six products are employed to collude snow cover. \cite{stigter2017assimilation}, has observed an accuracy of 83.1 \% with on-point snow compliances in the Langtang receptacle, the snow mapping algorithm utilizes MODIS bands 4 and 6 to calculate the regularized-difference snow indicator (\cite{hall2002modis}). In MOD10A2, a pixel is marked as snow if observed at least once in eight days, as no snow if absent in all eight days, and as pall if pall cover is observed every day. For this study, 496 images from 2002 to 2012 were obtained from the National Snow and Ice Data Center's (NSIDC) website. The 30 m-resolution ASTER Global Digital Elevation Model (GDEM) was sourced from \href{https://lpdaac.usgs.gov}{https://lpdaac.usgs.gov}. The study area boundary was outlined in the DEM. MOD10A2 images are projected to the World Geodetic System 1984 (WGS84), UTM Zone 45. The 8-day maximum SCA was extracted from projected snow images using the delineated boundary. Images with more than 10\% cloud cover were discarded. Finally, the daily SCA for 365 days in a year was interpolated or extrapolated from the 8-day maximum SCA using the cubical spline method.

\subsection{Dataset Preparation}
SCA, Q, T, and P are used as inputs for forecasting the snowmelt runoff. MinMaxScaler scales the data without losing its distribution for faster convergence during the training process. MinMaxScaler uses data value and subtracts it from the minimum value of the overall features which is then divided by the difference of the maximum value and the minimum value of features. We have used two days' data to predict the third day's SCA i.e. window size of 2. Furthermore, the nested CV of the model is carried out to check the generability of any unseen data with five folds for the outer loop and three folds for the inner loop. The tensor shape for our models is (window size, number of features) i.e. (2,3) for three inputs and (2,4) for four inputs. Keras' K-Fold is used for splitting the dataset with a random state of 42 with a splitting ratio of 80\% i.e. 3212 samples and 20\% i.e. 803 samples for training and testing respectively.

\begin{equation}
    MinMax Scaler (m)= \frac{x - xmin}{xmax - xmin},
\end{equation}

where,

x is the input value,

xmin is the minimum value of the column,

xmax is the maximum value of the column

\subsection{Experimental setup}
The simulations are performed for four different architectures:
\begin{enumerate}
    \item SVR
    \item LSTM
    \item Transformer
    \item TCN
\end{enumerate}
Each of the four different architectures was evaluated for two different inputs, M1 with inputs (T, Q, SCA, and P) and M2 with (T, Q, and SCA), respectively. The list of hyper-parameters used for the experiments are given in the table \ref{tab:hyper-parameter_table}.

Performance of all the four architectures are evaluated using the following metrics:
\begin{enumerate}
    \item KGE: KGE provides an overview of bias ratio, correlation, and variability. The value of KGE equal to 1 indicates a perfect agreement between simulations and observations.
        \begin{equation}
            KGE = 1 - \sqrt{({r - 1})^2 + ({\alpha}- 1)^2 + ({\beta}-1)^2},
        \end{equation}
    where, 
    
    $\quad \beta = \frac{\bar{Q'}}{\bar{Q}}$ is the bias ratio,
    
    $\bar{Q}$ is the average observed discharge,
    
    $\bar{Q'}$ is the average simulated discharge,
    
    $ \gamma = \frac{CV_s}{CV_o}$ is the variability,
    
    $CV_o$ is the observed coefficient of variation,
    
    $CV_s$ is the simulated coefficient of variation, and
    
    r is Pearson’s correlation coefficient.

    The optimum value of KGE r, ${\beta}$, and ${\gamma}$ is 1. No cross-correlation between the bias ratio and variability is ensured by using CV in calculating ${\gamma}$  (\cite{kling2012runoff}).
    
    \item NSE: The magnitude of residual variance concerning the observed data variance is provided by the NSE.
        \begin{equation}
            NSE = 1 - \frac{\sum_{t=1}^{N}(Q_t - Q_t')^2}{\sum_{t=1}^{N}(Q_t - \bar{Q})^2},
        \end{equation}
    where,
    
    ${Q_t}$ is the observed  discharge at time \textit{t},
    
    ${Q_t'}$ is the simulated  discharge at time \textit{t}, and
    
    ${\bar{Q}}$ denotes average observed discharge.
    
    \item $R^{2}$: The coefficient of determination (${R^2}$) provides the intensity of the relation between the measured and the predicted values. Its value ranges from 0 to 1, closer to 0 is a low correlation whereas, 1 represents a high correlation.
         \begin{equation}
            R^{2}=\left[ \frac{\sum_{t=1}^{n} (Q'_t - \bar{Q'}) (Q_t - \bar{Q})}{\sqrt{\sum_{t=1}^{n} (Q'_t - \bar{Q'})^2} \sqrt{\sum_{t=1}^{n} (Q_t - \bar{Q})^2}} \right]^2,
        \end{equation}
    where

    ${Q_t}$ is the observed  discharge at time \textit{t},
    
    ${Q_t'}$ is the simulated  discharge at time \textit{t},

    $\bar{Q}$ is the average observed discharge, and
    
    $\bar{Q'}$ is the average simulated discharge.
        
    \item RMSE: RMSE is an evaluation metric used to measure the average differences between predicted and actual values. It represents the standard deviation of the error, with a lower RMSE value indicating a better fit. RMSE is sensitive to large errors and is calculated using the following equation:
        \begin{equation}
            RMSE=\sqrt{\frac{\sum_{t=1}^{n} (Q'_t - Q_t)^2}{n}},
        \end{equation}
     where
    
    ${Q_t}$ is the observed  discharge at time \textit{t}, and
    
    ${Q_t'}$ is the simulated discharge at time \textit{t}.
  
    \item MAE: MAE is the absolute difference between measured and predicted values. A lower MAE represents lower error. The equation of MAE is given below:
        \begin{equation}
            MAE = \frac{1}{n} \sum_{t=1}^{n} |Q_t - Q'_t|,
        \end{equation}
    where,
    
    ${Q_t}$ is the observed  discharge at time \textit{t}, and
    
    ${Q_t'}$ is the simulated discharge at time \textit{t}.

\end{enumerate}

\begin{table}[h]
\centering
\caption{Hyper-parameters to tune four different models for optimal performance}
\label{tab:hyper-parameter_table}
\begin{tabular}{cll}
\hline
 Model & \multicolumn{1}{c}{Hyper-parameter}                                                                                                                                      & \multicolumn{1}{c}{Values}                                                                                                                                                                                                                                               \\ \hline
SVR            & \begin{tabular}[c]{@{}l@{}}\\C\\ Epsilon\\ Kernel\end{tabular}                                                                                                                      & \begin{tabular}[c]{@{}l@{}}\\{[}0.1, 1, 10{]}\\ {[}0.01, 0.1, 0.2{]}\\ {[}Linear, RBF{]}\end{tabular}                                                                                                                                                                               \\ \hline
LSTM           & \begin{tabular}[c]{@{}l@{}}\\LSTM Layers\\ LSTM Units\\ Dropout Rate\\ Optimizer\\ Learning Rate\end{tabular}                                                                       & \begin{tabular}[c]{@{}l@{}}\\{[}1, 2, 3{]}\\ {[}32, 64, 128{]}\\ {[}0.2, 0.3,..., 0.5{]}\\ {[}Adam, Adamax, RMSProp, SGD{]}\\ {[}0.0001, 0.001,..., 0.1{]}\end{tabular}                                                                                                             \\ \hline
Transformer    & \begin{tabular}[c]{@{}l@{}}\\Transformer Blocks\\ Head Size\\ Num Heads\\ FF Dim\\ Dropout Rate\\ Num MLP Layers\\ MLP Units\\ MLP Dropout\\ Optimizer\\ Learning Rate\end{tabular} & \begin{tabular}[c]{@{}l@{}}\\{[}2, 4, 6, 8{]}\\ {[}8, 16,..., 256{]}\\ {[}2, 4,..., 16{]}\\ {[}4, 8,..., 64{]}\\ {[}0.1, 0.2,..., 0.6{]}\\ {[}1, 2, 3{]}\\ {[}32, 64,..., 256{]}\\ {[}0.1, 0.2,..., 0.6{]}\\ Adam, Adamax, RMSProp, SGD\\ {[}0.0001, 0.001,..., 0.1{]}\end{tabular} \\ \hline
TCN            & \begin{tabular}[c]{@{}l@{}}\\TCN Layers\\ Num Filters\\ Kernel Size\\ Optimizer\\ Learning Rate\end{tabular}                                                                        & \begin{tabular}[c]{@{}l@{}}\\{[}1, 2, 3{]}\\ {[}32, 64,..., 128{]}\\ {[}2, 3, 4{]}\\ {[}Adam, Adamax, RMSProp, SGD{]}\\ {[}0.0001, 0.001,..., 0.1{]}\end{tabular}                                                                                                                   \\ \hline
\end{tabular}
\end{table}

\section{Methodology}
In this section, the proposed methodology for snowmelt runoff prediction is shown in Figure 5. At first, we preprocess the dataset using MinMaxScaler to scale the data and set the value of Windows size 2 for the third day’s SCA prediction. The prepared data are then split into outer and inner folds in a random state of 42 with a splitting ratio of 80\% i.e. 3212 samples and 20\% i.e. 803 samples for training and testing respectively. The split data are used for the traditional ML and DL architectures as shown in Figure 5. The trained models are evaluated using metrics such as RSME, MAE, $R^{2}$, KGE, and NSE which are used for the models' comparison. The performance metrics help to create the baseline for model performance comparison and choose the best-performing traditional ML or DL model.
\begin{figure}[h]
    \centering
    \includegraphics[width=14cm,height=6cm]{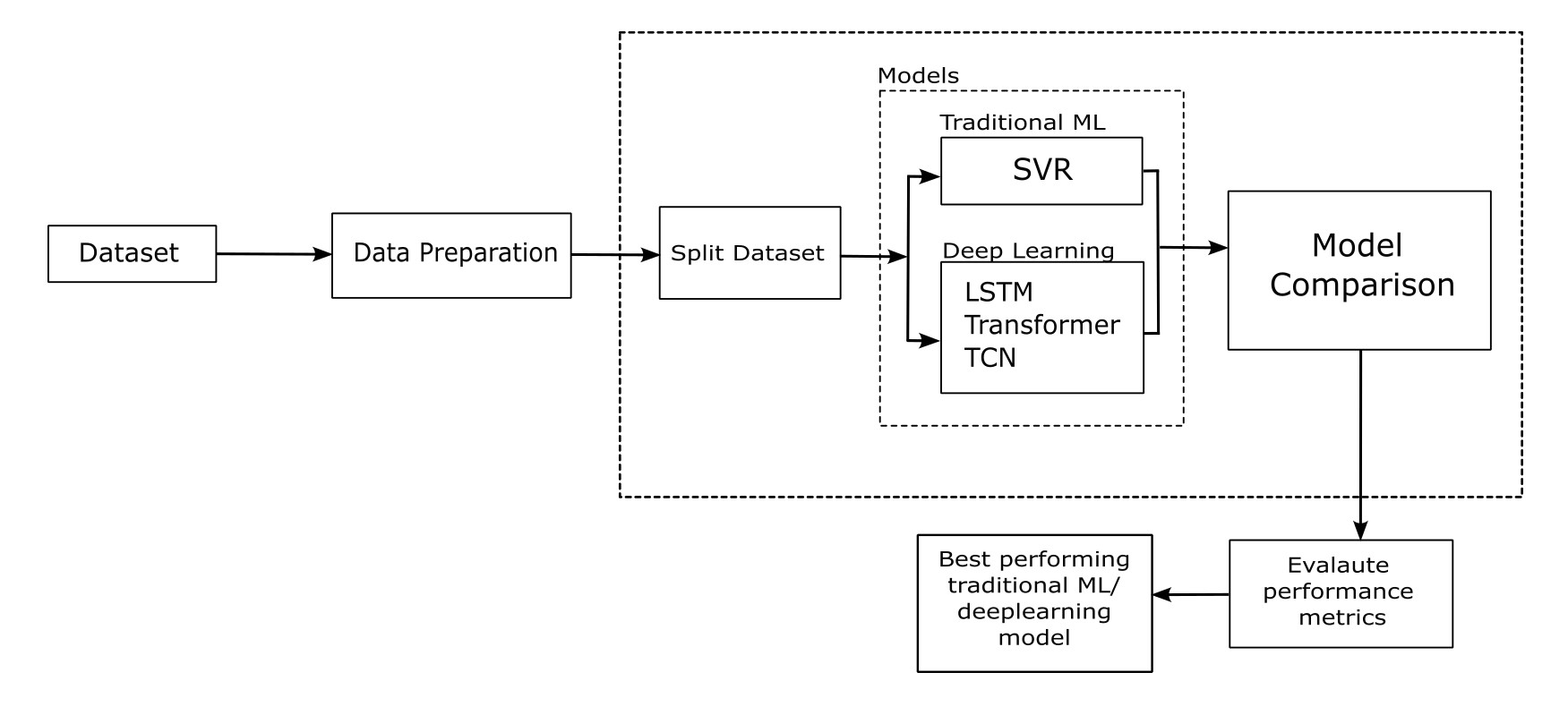} 
    \caption{Methodological framework for assessment of performance of ML techniques for snowmelt forecasting}
    \label{fig:5}
\end{figure}
\\In the figure \ref{fig:5}, the model section contains traditional ML and deep learning architecture. The detailed explanations of four different models used in the simulation are described below:
\subsection{SVR}
The SVR, proposed by \cite{vapnik1999nature} is a traditional ML algorithm for classification and regression. SVR can use both linear and non-linear kernels. SVR determines the hyperplane in high-dimensional space to separate the classes or output values. The main objective of SVR is to minimize the error value between the predicted and actual value for the correct prediction. In this study, we have used a Radial Basic Function (RBF) kernel with SVR, the RBF kernel is capable of capturing non-linear patterns. The rbf kernel mathematical expression is given below.

\begin{equation}
\
K(x_i, x_j) = \exp\left(-\frac{\|x_i - x_j\|^2}{2\sigma^2}\right),
\
\end{equation}
where,

\( x_i \) and \( x_j \) are the input feature vectors, and 

\( \sigma \) is a parameter controlling the width of the kernel.

In this study, we have used 0.1, 1, and 10 as regularization parameters (C), and 0.01, 0.1, and 0.2 are epsilon (\(\epsilon\)) values used for the hyper-parameter tuning. We have also used both linear and RBF in the kernel, all of those hyper-parameters are determined by the grid search approach. The best hyperparameters after tuning are 0.1 for C, 0.01 for epsilon, and linear kernel.

\subsection{LSTM}
\begin{figure}[h]
    \centering
    \includegraphics[width=15cm]{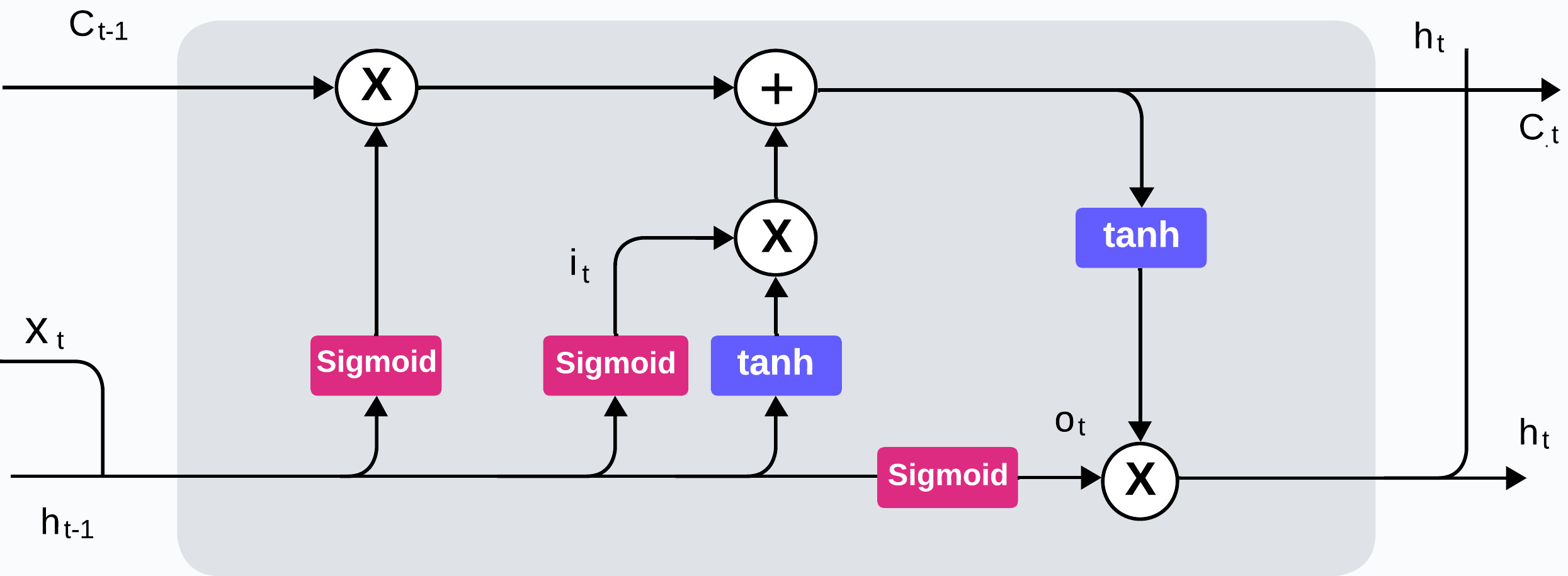} 
    \caption{A classic LSTM cell, where \textit{sigmoid} denotes sigmoidal function, \textit{tanh} denotes a hyperbolic tangent function, \textit{C} denotes cell state, \textit{h} denotes hidden state, \textit{o} denote output gate, \textit{i} denote input vectors at time step \textit{t}}
    \label{fig:2}
\end{figure}

LSTM proposed by \cite{hochreiter1997long} is a variation of RNN used widely in DL  architecture. Unlike the vanilla RNN, the LSTM is capable of capturing the long-term dependencies and solving the problem of exploding and vanishing gradient problems. LSTM consists of feedback connections, which allow it to process sequences of data. It consists of three different gates: the input gate, the forget gate, and the output gate. The forget gate removes the information that is no longer useful in the cell state. Additional new information is added in the cell state by the input gate. The output gate extracts the useful information from the cell state to be presented as output. A classic LSTM cell is shown in figure \ref{fig:2} and its related equations are below.

Forget gate: \( f_t = \sigma(w_f \cdot [h_{t-1}, x_t] + b_f) \)

Input gate: \( i_t = \sigma(w_i \cdot [h_{t-1}, x_t] + b_i) \)

Update vector: \( \hat{c} = \tanh(w_c \cdot [h_{t-1}, x_t] + b_c) \)

Cell state: \( f_t \odot C_{t-1} + i_t \odot \hat{c}_t \)

Output gate: \( \sigma(w_o \cdot [h_{t-1}, x_t] + b_o) \)

where,

\( w_f \), \( w_i \), \( w_c \), and \( w_o \) are weights for forget gate, input gate, cell state, and output gate respectively,

\( b_f \), \( b_i \), \( b_c \), and \( b_o \) are biases for the forget gate, input gate, cell state, and output gate respectively,

\( \sigma \) represents the sigmoid activation function, and

\([h_{t-1}, x_t]\) represents the concatenation of the current input and the previous hidden state.
The hyper-parameters used in the LSTM model are three different numbers of LSTM layers, three different LSTM hidden units with values 32, 64, and 128, dropout values starting from 0.2 to 0.5, learning rate values starting from 0.0001 to 0.1, and five different optimizers such as Adam, Adamax, RMSProp, and SGD are used for tuning model. The hyper-parameter tuning is done using a random search in the Keras tuner. For the optimal hyperparameters in our model, we utilized an Adam optimizer with a learning rate of 0.004, dropout rate of 0.2, and incorporated two LSTM layers each consisting of 96 hidden units.

\subsection{Transformer}
\begin{figure}[h]
    \centering
    \includegraphics[width=12cm,height=16cm]{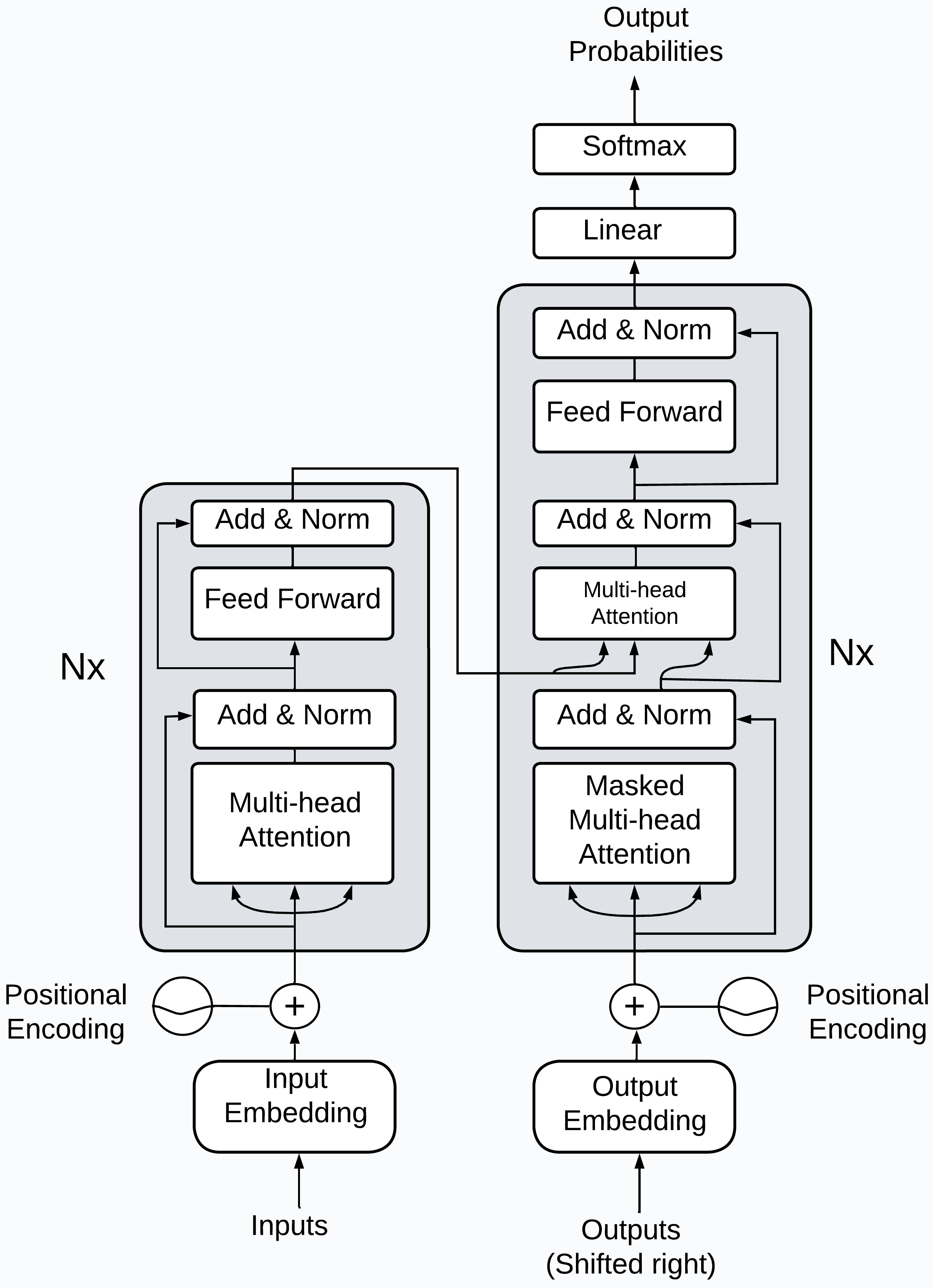} 
    \caption{The detail model architecture of the Transformer. [\cite{vaswani2017attention}]}
    \label{fig:3}
\end{figure}

A transformer proposed by \cite{vaswani2017attention} is a DL architecture, that includes an encoder and decoder. These encoder and decoder are connected through the attention mechanism. This architecture requires less training time than LSTM, as its latest version is highly used for training large language models (LLM). This architecture is applicable for natural language processing (NLP) and computer vision, but also in audio and multi-modal processing. Figure \ref{fig:3} shows transformer model architecture and its components are described below.

\begin{enumerate}
    \item Self-Attention Mechanism: The self-attention mechanism is responsible for capturing any crucial information from the long sequence of data. The calculation for attention can be expressed as a large matrix calculation using the softmax function.
    \begin{equation}
        \text{Attention}(\mathbf{Q, K, V}) = \text{softmax}\left(\frac{\mathbf{QK}^T}{\sqrt{d_k}}\right)\mathbf{V},
    \end{equation}
where,

\( \mathbf{Q} \) represents Query,

\( \mathbf{K} \) represents key, 

\( \mathbf{V} \) represents value, and 

\( d_k \) represents the dimensionality of the key in items 
    \item Multi-Head Attention: The multi-head attention is a feature of the transformer that helps the model process different aspects of the input sequence simultaneously. In this process, multiple self-attention heads work in parallel, and outputs are concatenated and linearly transformed.
        \begin{equation}
        \text{MultiHead}(\mathbf{Q, K, V}) = \text{Concat}(\mathbf{h}_1, \ldots, \mathbf{h}_n) \mathbf{W}_o,
        \end{equation}
    where,
    
    \( \mathbf{h} \) represents the heads, and 
    
    \( \mathbf{W}_o \) represents the matrix of the entire multi-head attention.
    \item Position Encoding: Position encoding in the architecture is responsible for the order of the words in a sequence, and this is added to the input embedding.
        \begin{equation}
            \text{PE}(\text{pos}, 2i) = \sin\left(\frac{\text{pos}}{1000^{(2i/d_{\text{model}})}}\right),   
        \end{equation}
        \begin{equation}
            \text{PE}(\text{pos}, 2i+1) = \cos\left(\frac{\text{pos}}{1000^{(2i/d_{\text{model}})}}\right),
        \end{equation}
    where,
    
    pos is the position of the sequence, and
    
    \(d_{\text{model}}\) is the dimensionality of the model.
\end{enumerate}

In the transformer, we use four different transformer blocks with values 2,4,6, and 8, head size starting from 8 to 256, number of heads from 2 to 16, dropout rate with values starting from 0.1 to 0.6, three different numbers of multi-layer perceptron (MLP) layers, MLP units values from 32 to 256, MLP Dropout values starting from 0.1 to 0.6, four different optimizers such as Adam, SGD, RMSProp, and Adamax, and learning rate with values from 0.0001 to 0.1 are used as hyper-parameters to tune transformer model. We have used a Keras tuner for the hyper-parameter tuning. The optimal hyperparameters are two transformer blocks, each with a head size of 136, and two attention heads. The two MLP layers were employed with units of 192 and 160, respectively. Dropout regularization was applied across the model, with rates of 0.41 for the first layer, 0.10 for the second layer, and 0.11 for the subsequent layers. The Adam optimizer with a learning rate of 0.0044 was used for training.

\subsection{TCN}
The TCN proposed by \cite{bai2018empirical} is a DL architecture that is capable of capturing long-term dependencies and temporal patterns. TCN architecture is capable of preventing data leakage using causal convolution by ensuring that the model's prediction is solely on past and present information. The TCN architecture can handle the sequence of varying lengths, allowing for efficient capture of long-range dependencies. Figure \ref{fig:4} shows the TCN model architecture and its key components are described below.
\begin{enumerate}
    \item Dilated Causal Convolutions: In TCN, the dilated causal convolutions can process any length of sequence at varying receptive field sizes. The dilated convolution is responsible for capturing the long dependencies and is useful for memory management.
    \begin{equation}
        y_t = \sum_{i=1}^k w_i \cdot x_{t-d_t},
    \end{equation}    
    where,
    
    \(y_t\) is the output at time \(t\),
    
    \(w_i\) is the weight of the convolutional filter,
    
    \(x_t\) is the input at time \(t\),
    
    \(d\) is the dilation rate, and
    
    \(k\) is the filter size.
    \item Residual Blocks: TCN uses residual connections within its blocks which help to solve the vanishing gradient problem and allow for effective optimization of deeper architecture. TCN’s residual block consists of two layers of dilated causal convolution and non-linearity, for which a rectified linear unit is being used. 
    \item Temporal Skip connection: TCN has skip connections that connect every other layer, which makes the network capable of capturing and reusing the information. The skip connection is also useful for promoting gradient flow also can enhance the model's ability to learn hierarchical representation.
    \begin{figure}[h]
        \centering
        \includegraphics[width=16.5cm,height=8cm]{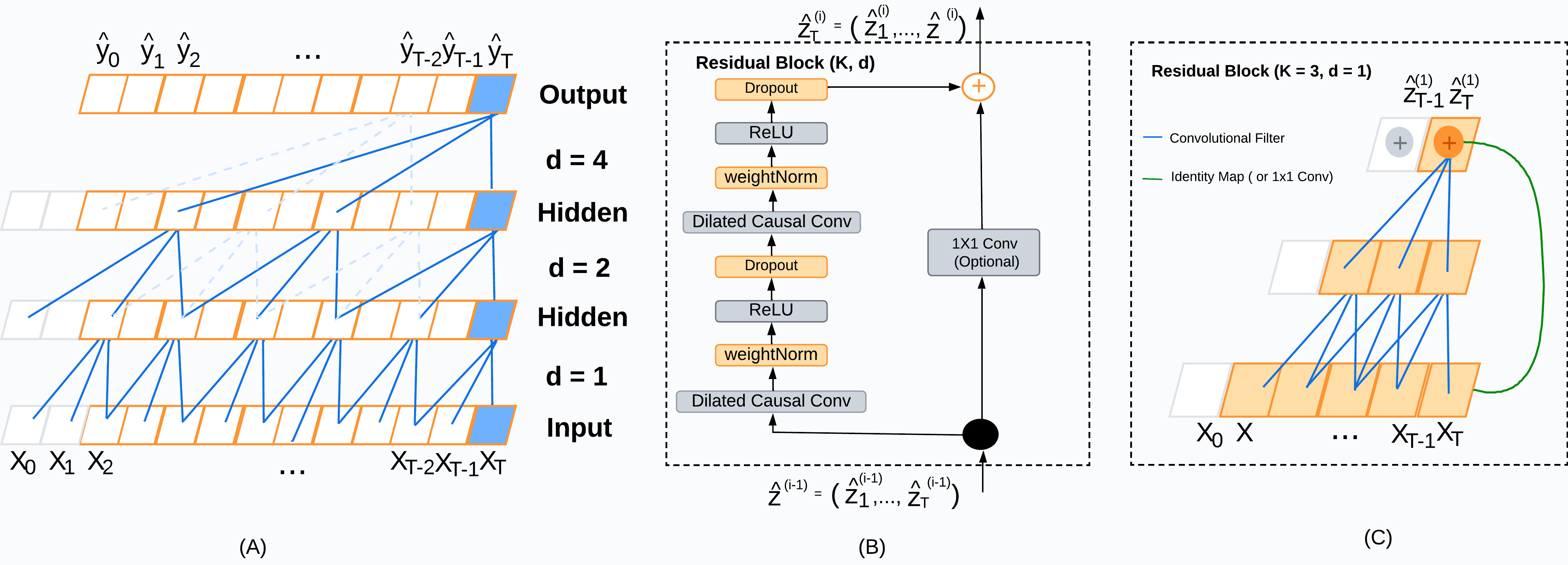} 
        \caption{A TCN architectural elements. (A) A dilated causal convolution with dilated factors d = 1, 2, 4 and filter size k = 3.  (B) TCN residual block. A 1 \,x\, 1 convolution matrix is added when residual input and output have different dimensions. (C) A residual connection example is in a TCN where the blue line represents residual function and the green lines represent identity mappings. (\cite{bai2018empirical})}
        \label{fig:4}
    \end{figure}
\end{enumerate}
In this study, the TCN model uses three different TCN layers, we used 32 to 256 filters, three kernel sizes such as 2,3, and 4, five different optimizers like Adam, Adamax, RMSProp, and SGD, and learning rates values starting from 0.0001 to 0.1 use as hyper-parameters for tuning TCN model. we have used a keras tuner for hyper-parameter tuning.  The optimal hyperparameters are a single layer of TCN with 96 filters, and a kernel size of 3 with Adam as optimizer along a learning rate of 0.0057.

\section{Results}
This section presents a comprehensive analysis of the model's performance across different scenarios. We evaluate the effectiveness of both three-input and four-input models in each fold, comparing their performance against other models. Additionally, we provide insights into the testing time required for each model.
\subsection{Performance analysis of ML models with four inputs}
This subsection describes the performance of ML models with four inputs in each fold of nested CV as given in table \ref{tab:inputs_4}.

Table \ref{tab:inputs_4} shows the performance of the DL architectures and the traditional ML algorithms. SVR model shows the least performance for each fold of nested CV among all other models with average MAE, RMSE, ${R^2}$, KGE, and NSE values of 0.032, 0.042, 0.97, 0.918, and 0.97, respectively. The LSTM model shows a significant performance improvement over SVR in terms of average MAE, RMSE, ${R^2}$, KGE, and NSE of 0.013, 0.025, 0.989, 0.975, and 0.989, respectively. The transformer achieves better performance as compared to LSTM with an average KGE value of 0.986. However, TCN shows superior performance as compared to other models due to its capability of capturing long-range dependencies where average performance metrics values of MAE, RMSE, ${R^2}$, KGE, and NSE are 0.011, 0.023, 0.991, 0.992, and 0.991, respectively. It's important to note that we used data points from the testing set of the best-predicted model fold to avoid data leakage resulting from systematic data separation using nested CV. In the scatter plot diagram, the points that lie near the diagonal line (1:1 line), estimate the river discharge accurately, whereas points above and below the diagonal overestimate and underestimate the actual flow, respectively. Figure \ref{fig:6} shows that the SVR and transformer model predicts good for the medium flow but the higher and lower are overestimated and underestimated, respectively. The LSTM model predicts good for low and medium flows but the high flows are underestimated by the model except the TCN model. Therefore, the TCN model predicts peak flows accurately as compared to other models.

\begin{figure}[h]
    \centering
    \includegraphics[width=16cm]{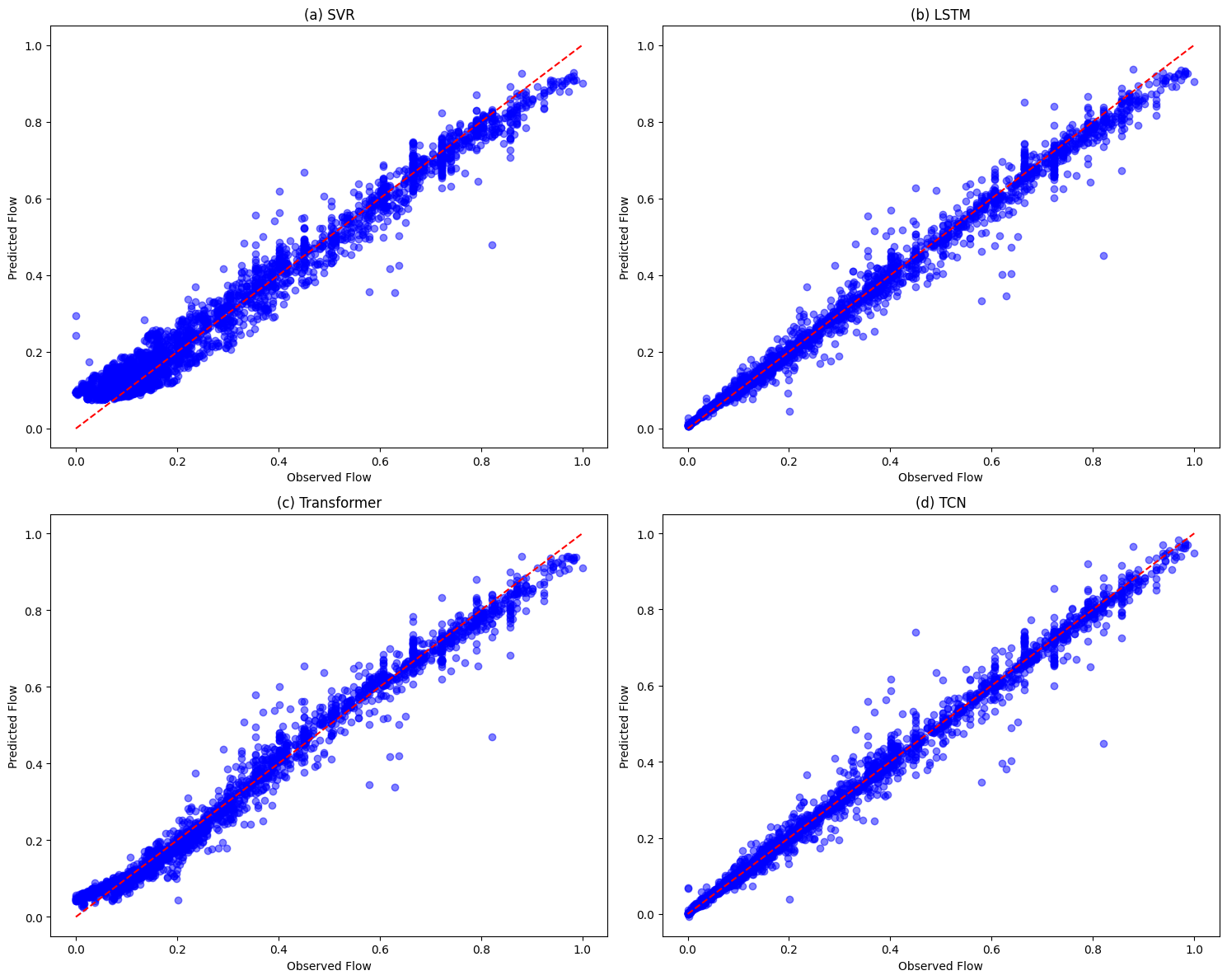} 
    \caption{Scatter plot of predicted flow with observed flow (a) SVR model (b) LSTM model (c) Transformer model (d) TCN model. The solid line is the 1:1 line. }
    \label{fig:6}
\end{figure}

\begin{table}[]
  \centering
    \caption{Performance Comparison of different models with four inputs in each folds by using MAE, RMSE, ${R^2}$, KGE, and NSE}.
    \label{tab:inputs_4}
\begin{tabular}{ccccccc}
\hline
\multicolumn{1}{l}{Models} & \multicolumn{1}{l}{Folds}                                             & \multicolumn{1}{l}{MAE}                                                                        & \multicolumn{1}{l}{RMSE}                                                                       & \multicolumn{1}{l}{${R^2}$}                                                                     & \multicolumn{1}{l}{KGE}                                                                        & \multicolumn{1}{l}{NSE}                                                                        \\ \hline
\\SVR                        & \begin{tabular}[c]{@{}c@{}}1\\ \\ 2\\ \\ 3\\ 
\\ 4\\ \\ 5\end{tabular} & \begin{tabular}[c]{@{}c@{}}0.0341\\ \\ 0.0312\\ \\ 0.0306\\ \\ 0.0338\\ \\ \textbf{0.0313}\end{tabular}
& \begin{tabular}[c]{@{}c@{}}0.0451\\ \\ 0.0401\\ \\ 0.0406\\ \\ 0.0450\\ \\ \textbf{0.0395}\end{tabular}
& \begin{tabular}[c]{@{}c@{}}0.9657\\ \\ 0.9723\\ \\ 0.9710\\ \\ 0.9672\\ \\ \textbf{0.9734}\end{tabular}
& \begin{tabular}[c]{@{}c@{}}0.8967\\ \\ 0.9291.\\ \\ 0.9227\\ \\ 0.9175\\ \\ \textbf{0.9220}\end{tabular}
& \begin{tabular}[c]{@{}c@{}}0.9657\\ \\ 0.9723\\ \\ 0.9710\\ \\ 0.9672\\ \\ \textbf{0.9734}\end{tabular} \\ \hline
\\LSTM                       
& \begin{tabular}[c]{@{}c@{}}1\\ \\ 2\\ \\ 3\\ \\ 4\\ \\ 5\end{tabular} & \begin{tabular}[c]{@{}c@{}}0.0161\\ \\ 0.0130\\ \\ 0.0126\\ \\ 0.0130\\ \\ \textbf{0.0121}\end{tabular} 
& \begin{tabular}[c]{@{}c@{}}0.0301\\ \\ 0.0235\\ \\ 0.0239\\ \\ 0.0242\\ \\ \textbf{0.0220}\end{tabular} 
& \begin{tabular}[c]{@{}c@{}}0.9847\\ \\ 0.9905\\ \\ 0.9901\\ \\ 0.9905\\ \\ \textbf{0.9918}\end{tabular}
& \begin{tabular}[c]{@{}c@{}}0.9727\\ \\ 0.9889\\ \\ 0.9693\\ \\ 0.9713\\ \\ \textbf{0.9735}\end{tabular}
& \begin{tabular}[c]{@{}c@{}}0.9847\\ \\ 0.9905\\ \\ 0.9903\\ \\ 0.9905\\ \\ \textbf{0.9918}\end{tabular} \\ \hline
\\Transformer                
& \begin{tabular}[c]{@{}c@{}}1\\ \\ 2\\ \\ 3\\ \\ 4\\ \\ 5\end{tabular} 
& \begin{tabular}[c]{@{}c@{}}0.0195\\ \\ 0.0177\\ \\ 0.0186\\ \\ 0.0185\\ \\ \textbf{0.0177}\end{tabular} 
& \begin{tabular}[c]{@{}c@{}}0.0308\\ \\ 0.0274\\ \\ 0.0283\\ \\ 0.0281\\ \\ \textbf{0.0258}\end{tabular} 
& \begin{tabular}[c]{@{}c@{}}0.9840\\ \\ 0.9870\\ \\ 0.9860\\ \\ 0.9872\\ \\ \textbf{0.9887}\end{tabular}  
& \begin{tabular}[c]{@{}c@{}}0.9855\\ \\ 0.9880\\ \\ 0.9819\\ \\ 0.9859\\ \\ \textbf{0.9883}\end{tabular} 
& \begin{tabular}[c]{@{}c@{}}0.9840\\ \\ 0.9870\\ \\ 0.9860\\ \\ 0.9872\\ \\ \textbf{0.9887}\end{tabular} \\ \hline
\\TCN                        
& \begin{tabular}[c]{@{}c@{}}1\\ \\ 2\\ \\ 3\\ \\ 4\\ \\ 5\end{tabular} 
& \begin{tabular}[c]{@{}c@{}}0.0126\\ \\ 0.0110\\ \\ 0.0107\\ \\ 0.0109\\ \\ \textbf{0.0098}\end{tabular} 
& \begin{tabular}[c]{@{}c@{}}0.0269\\ \\ 0.0227\\ \\ 0.0214\\ \\ 0.0230\\ \\ \textbf{0.0192}\end{tabular} 
& \begin{tabular}[c]{@{}c@{}}0.9878\\ \\ 0.9911\\ \\ 0.9919\\ \\ 0.9914\\ \\ \textbf{0.9937}\end{tabular} 
& \begin{tabular}[c]{@{}c@{}}0.9914\\ \\ 0.9927\\ \\ 0.9913\\ \\ 0.9926\\ \\ \textbf{0.9931}\end{tabular} 
& \begin{tabular}[c]{@{}c@{}}0.9878\\ \\ 0.9911\\ \\ 0.9919\\ \\ 0.9914\\ \\ \textbf{0.9937}\end{tabular} \\ \hline
\end{tabular}
\end{table}

\subsection{Performance analysis of ML models with three inputs}
In this subsection, the performance of ML models with three inputs in each fold of nested CV is given in the table \ref{tab:inputs_3}.

SOTA DL architectures achieve higher performance as compared to the traditional  ML algorithms as shown in table \ref{tab:inputs_3}. SVR models have the minimum performance over each fold in comparison to other models with average MAE, RMSE, ${R^2}$, KGE, and NSE values of 0.030, 0.039, 0.974, 0.937, and 0.974, respectively. The LSTM model shows a significant performance improvement over SVR with average MAE, RMSE, ${R^2}$, KGE, and NSE of 0.012, 0.024, 0.99, 0.981, and 0.99, respectively. The LSTM shows better performance with an average KGE value of 0.981 than the transformer model with an average KGE value of 0.977. However, TCN shows superior performance also for the 3 inputs due to its capability of capturing long-range dependencies where average performance metrics values of MAE, RMSE, ${R^2}$, KGE, and NSE are 0.011, 0.023, 0.991, 0.987, and 0.991, respectively. Figure \ref{fig:6} shows that the models predict good for low and medium flows but the high flows are underestimated by all models except the TCN model.

\begin{table}[h]
  \centering
    \caption{Performance Comparison of different models with three inputs in each folds by using MAE, RMSE, ${R^2}$, KGE, and NSE}.
    \label{tab:inputs_3}
\begin{tabular}{ccccccc}
\hline
\multicolumn{1}{l}{Models} & \multicolumn{1}{l}{Folds}                                                & \multicolumn{1}{l}{MAE}                                                                        & \multicolumn{1}{l}{RMSE}                                                                       & \multicolumn{1}{l}{${R^2}$}                                                                    & \multicolumn{1}{l}{KGE}                                                                        & \multicolumn{1}{l}{NSE}                                                                        \\ \hline
\\SVR                        
& \begin{tabular}[c]{@{}c@{}}1\\ \\ 2\\ \\ 3\\ \\ 4\\ \\ 5\end{tabular} & \begin{tabular}[c]{@{}c@{}}0.0317\\ \\ 0.0300\\ \\ 0.0288\\ \\ 0.0300\\ \\ \textbf{0.0284}\end{tabular} 
& \begin{tabular}[c]{@{}c@{}}0.0426\\ \\ 0.0386\\ \\ 0.0374\\ \\ 0.0395\\ \\ \textbf{0.0368}\end{tabular} 
& \begin{tabular}[c]{@{}c@{}}0.9693\\ \\ 0.9743\\ \\ 0.9755\\ \\ 0.9747\\ \\ 
\textbf{0.9769}\end{tabular} 
& \begin{tabular}[c]{@{}c@{}}0.9316\\ \\ 0.9403\\ \\ 0.9405\\ \\ 0.9335\\ \\ \textbf{0.9369}\end{tabular} 
& \begin{tabular}[c]{@{}c@{}}0.9693\\ \\ 0.9743\\ \\ 0.9755\\ \\ 0.9746\\ \\ \textbf{0.9769}\end{tabular} \\ \hline
\\LSTM                       
& \begin{tabular}[c]{@{}c@{}}1\\ \\ 2\\ \\ 3\\ \\ 4\\ \\ 5\end{tabular}    & \begin{tabular}[c]{@{}c@{}}0.0143\\ \\ 0.0136\\ \\ 0.0112\\ \\ 0.0116\\ \\ \textbf{0.0106}\end{tabular} & 
\begin{tabular}[c]{@{}c@{}}0.0272\\ \\ 0.0236\\ \\ 0.0231\\ \\ 0.0232\\ \\ \textbf{0.0206}\end{tabular} & 
\begin{tabular}[c]{@{}c@{}}0.9875\\ \\ 0.9903\\ \\ 0.9905\\ \\ 0.9912\\ \\ \textbf{0.9927}\end{tabular} & 
\begin{tabular}[c]{@{}c@{}}0.9768\\ \\ 0.9757\\ \\ 0.9812\\ \\ 0.9840\\ \\ \textbf{0.9875}\end{tabular} & 
\begin{tabular}[c]{@{}c@{}}0.9875\\ \\ 0.9903\\ \\ 0.9905\\ \\ 0.9912\\ \\ \textbf{0.9927}\end{tabular} \\ \hline
\\Transformer                
& \begin{tabular}[c]{@{}c@{}}1\\ \\ 2\\ \\ 3\\ \\ 4\\ \\ 5\end{tabular}    & \begin{tabular}[c]{@{}c@{}}0.0178\\ \\ 0.0161\\ \\ 0.0162\\ \\ 0.0126\\ \\ \textbf{0.0115}\end{tabular} & 
\begin{tabular}[c]{@{}c@{}}0.0305\\ \\ 0.0262\\ \\ 0.0266\\ \\ 0.0243\\ \\ \textbf{0.0213}\end{tabular} & 
\begin{tabular}[c]{@{}c@{}}0.9842\\ \\ 0.9881\\ \\ 0.9875\\ \\ 0.9904\\ \\ \textbf{0.9922}\end{tabular} & 
\begin{tabular}[c]{@{}c@{}}0.9663\\ \\ 0.9781\\ \\ 0.9751\\ \\ 0.9826\\ \\ \textbf{0.9831}\end{tabular} & 
\begin{tabular}[c]{@{}c@{}}0.9842\\ \\ 0.9881\\ \\ 0.9875\\ \\ 0.9904\\ \\ \textbf{0.9922}\end{tabular} \\ \hline
\\TCN                        
& \begin{tabular}[c]{@{}c@{}}1\\ \\ 2\\ \\ 3\\ \\ 4\\ \\ 5\end{tabular}    & \begin{tabular}[c]{@{}c@{}}0.0122\\ \\ 0.0108\\ \\ 0.0113\\ \\ 0.0115\\ \\ \textbf{0.0104}\end{tabular} & 
\begin{tabular}[c]{@{}c@{}}0.0269\\ \\ 0.0217\\ \\ 0.0223\\ \\ 0.0228\\ \\ \textbf{0.0201}\end{tabular} & 
\begin{tabular}[c]{@{}c@{}}0.9877\\ \\ 0.9918\\ \\ 0.9913\\ \\ 0.9915\\ \\ \textbf{0.9931}\end{tabular} & 
\begin{tabular}[c]{@{}c@{}}0.9806\\ \\ 0.9891\\ \\ 0.9865\\ \\ 0.9888\\ \\ \textbf{0.9904}\end{tabular} & 
\begin{tabular}[c]{@{}c@{}}0.9877\\ \\ 0.9918\\ \\ 0.9913\\ \\ 0.9915\\ \\ \textbf{0.9931}\end{tabular} \\ \hline
\end{tabular}
\end{table}

\begin{figure}[h]
    \centering
    \includegraphics[width=16cm]{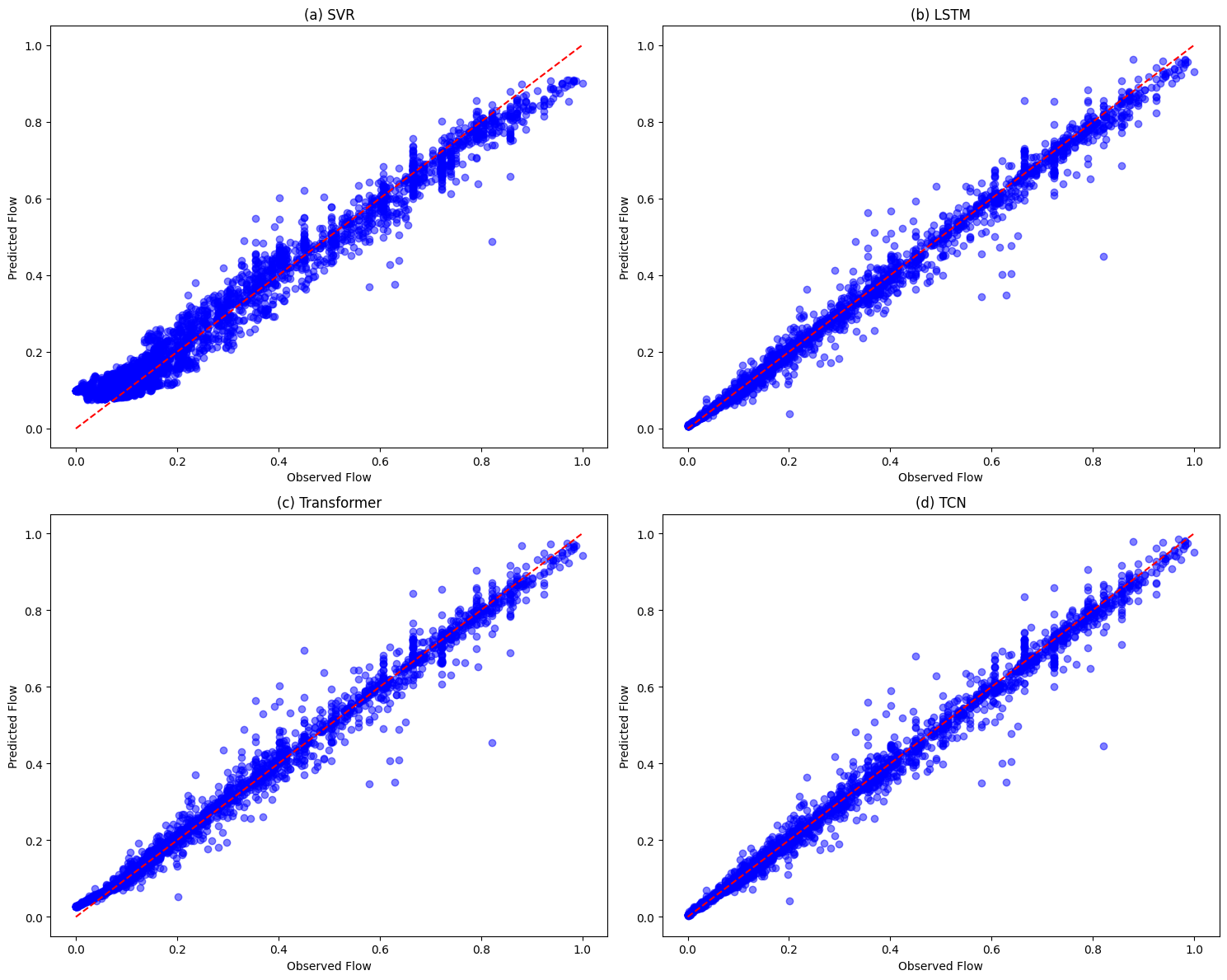} 
    \caption{Scatter plot of predicted flow with observed flow (a) SVR model (b) LSTM model (c) Transformer model (d) TCN model. The solid line is the 1:1 line. }
    \label{fig:6}
\end{figure}
\subsection{Overall comparison of ML models }
This section describes the overall performance of the models by averaging all the performance metrics for M1 and M2 input types given in the table \ref{tab:model_comparison}. 

Table \ref{tab:model_comparison} shows the value of different performance metrics obtained by all the models over five outer folds. The sensitivity analysis of the models for two input conditions, i.e., all four inputs (M1) and all inputs without pressure (M2), is presented in the table. The SVR model performs better for M2 input with average KGE = 93.7\% and RMSE = 0.039, as compared to M1 input with an average KGE and RMSE of 91.8\% and 0.042, respectively. Similarly, the LSTM model also performs better with M2 input with an average KGE of 98.1\% and RMSE of 0.024 and outperforms M1 input with an average KGE and RMSE of 97.5\% and 0.025, respectively. Although the transformer model shows better performance for all the other performance metrics in the M2 input except for the average KGE. Specifically, the value of the average KGE with M2 input is 97.7\%, and for the M1 input is 98.6\%. Finally, the table \ref{tab:model_comparison} concludes that the TCN model has better performance as compared to any other models. Unlike other models, TCN performance is better with M1 input with an average KGE of 99.2\%, which is 0.5\% greater than M2 on average. However, in this overall comparison section the results we have obtained in input M1, and M2 are almost similar. The result obtained in the study by \cite{thapa2020snowmelt} has a very close RMSE value for the M1 and M2 inputs. The final value of RMSE reached in the aforementioned paper is achieved using the hit and trial method which is questionable whether the RMSE could further be reduced, which could increase the model's performance. Hence our research contributes to the fact that the use of a hyper-parameter tuning can significantly reduce the RMSE value for both M1 and M2 as compared to the hit and trial method used in Thapa et al. Furthermore, the research contributes the addition of the P parameter has no significant contribution to snowmelt prediction.
\begin{table}[]
  \centering
    \caption{Overall performance Comparison with four inputs M1 and three inputs M2 for various models; where performance metrics are: MAE, RMSE, $R^{2}$, KGE, and NSE}.
  \label{tab:model_comparison}
\begin{tabular}{ccccccc}
\hline
Models      & INPUTS & MAE& RMSE & ${R^2}$ & KGE  & NSE                                                      \\ \hline
\\SVR         
& \begin{tabular}[c]{@{}c@{}}M1\\ \\ M2\end{tabular} & \begin{tabular}[c]{@{}c@{}}0.032\\ \\ \textbf{0.030}\end{tabular} 
& \begin{tabular}[c]{@{}c@{}}0.042\\ \\ \textbf{0.039}\end{tabular} & \begin{tabular}[c]{@{}c@{}}0.970\\ \\ \textbf{0.974}\end{tabular} & \begin{tabular}[c]{@{}c@{}}0.918\\ \\ \textbf{0.937}\end{tabular}                           & \begin{tabular}[c]{@{}c@{}}0.970\\ \\ \textbf{0.975}\end{tabular} \\ \hline
\\LSTM        & \begin{tabular}[c]{@{}c@{}}M1\\ \\ M2\end{tabular} & \begin{tabular}[c]{@{}c@{}}0.013\\ \\ \textbf{0.012}\end{tabular} & \begin{tabular}[c]{@{}c@{}}0.025\\ \\ \textbf{0.024}\end{tabular} & \begin{tabular}[c]{@{}c@{}}0.989\\ \\ \textbf{0.990}\end{tabular} & \begin{tabular}[c]{@{}c@{}}0.975\\ \\ \textbf{0.981}\end{tabular}                           & \begin{tabular}[c]{@{}c@{}}0.989\\ \\ \textbf{0.990}\end{tabular} \\ \hline
\\Transformer & \begin{tabular}[c]{@{}c@{}}M1\\ \\ M2\end{tabular} & \begin{tabular}[c]{@{}c@{}}0.018\\ \\ \textbf{0.015}\end{tabular} & \begin{tabular}[c]{@{}c@{}}0.028\\ \\ \textbf{0.026}\end{tabular} & \begin{tabular}[c]{@{}c@{}}0.987\\ \\ \textbf{0.989}\end{tabular} & \begin{tabular}[c]{@{}c@{}}\textbf{0.986}\\ \\ 0.977\end{tabular}                           & \begin{tabular}[c]{@{}c@{}}0.987\\ \\ \textbf{0.989}\end{tabular} \\ \hline
\\TCN         & \begin{tabular}[c]{@{}c@{}}M1\\ \\ M2\end{tabular} & \begin{tabular}[c]{@{}c@{}}\textbf{0.011}\\ \\ 0.011\end{tabular} & \begin{tabular}[c]{@{}c@{}}\textbf{0.023}\\ \\ 0.023\end{tabular} & \begin{tabular}[c]{@{}c@{}}\textbf{0.991}\\ \\ 0.991\end{tabular} & \begin{tabular}[c]{@{}c@{}}\textbf{0.992}\\ \\ 0.987\end{tabular} & \begin{tabular}[c]{@{}c@{}}\textbf{0.991}\\ \\ 0.991\end{tabular} \\ \hline
\end{tabular}
\end{table}

\subsection{Testing time}
In this subsection, we have computed the testing time in each outer fold for different models and averaged its testing time as shown in table \ref{tab:model_testing_time}. We have not reported the testing time of the SVR model based on several considerations, including the computational efficiency of SVR relative to other models evaluated in our study. in comparison to more complex models like LSTM, TCN, and Transformers. The training and testing times for SVR are generally lower due to its simpler architecture and optimization algorithms, which are specifically designed for efficiency in high-dimensional feature spaces. By focusing our testing time analysis on LSTM, TCN, and Transformers, we aimed to highlight the computational characteristics of these newer and more sophisticated models, which are of greater interest to the research community given their potential for handling complex sequential data. The table shows the TCN model has a faster computational testing time than any other DL architecture, as shown in table \ref{tab:model_testing_time}.

\begin{table}[h]
  \centering
    \caption{Model testing time (seconds) comparison}.
  \label{tab:model_testing_time}
\begin{tabular}{cl}
\hline
\\Models      & \multicolumn{1}{c}{Testing Time (seconds)} \\ \hline\\
LSTM        & 0.694                                      \\ \hline\\
Transformer & 0.652                                      \\ \hline\\
TCN         & \textbf{0.372}                                      \\ \hline
\end{tabular}
\end{table}

\section{Discussion}
Snowmelt modeling is useful for estimating the water availability for reservoir management, water supply, irrigation, and hydroelectricity projects, so it is necessary to explore new techniques for the appropriate snowmelt forecasting. Taking Langtang Basin as an example, we applied SOTA DL architectures such as Transformer and TCN for the snowmelt prediction which has not been used for the snowmelt forecasting. In this study, model efficiency (in terms of NSE) for SVR, LSTM, Transformer, and TCN were 97.5\%, 99\%, 98.9\%, and 99.1\%, respectively. In a previous study by \cite{uysal2016improving}, the ANN model was used for the snowmelt runoff prediction in the Upper Euphrates Basin of Turkey and achieved 93\% model efficiency which is lower than the ML model used in this study. In the study by \cite{le2019application}, the LSTM model has a model efficiency of 99.2\% comparable to our LSTM model result used in this study. 

\cite{kratzert2018rainfall} used two-layered LSTM whereas \cite{le2019application} and \cite{fan2020comparison} used single-layer LSTM, these studies did not evaluate the different numbers of LSTM layers. In a study by \cite{thapa2020snowmelt}, the performance of the LSTM model for the different hidden layers, optimizers, and window size was evaluated, however, the manual tuning of hyper-parameter without the use of any systematic approach is questionable for achieving the best performance of the model. In our study, we use a Keras tuner for the hyper-parameter tuning, so that each combination of the hyper-parameter will be evaluated to find the best prediction model. After hyper-parameter tuning, we found that two-layered stacked performs better than single-layer LSTM. In this study, we have compared the performance of five optimizers and found that Adam with a learning rate of 0.004 performs better. The performance of all the models for each fold is shown in Table \ref{tab:inputs_4} and \ref{tab:inputs_3}.

Most of the studies employing ML models (\cite{uysal2016improving, kratzert2018rainfall, fan2020comparison, 
 thapa2020snowmelt}) does not exploit the recent SOTA deep learning architectures. In the past endeavor, traditional approaches such as SVR, ANN, and LSTM were used for snowmelt forecasting. In our study, we have used recent DL architectures such as Transformer and TCN, these models have claimed that they are capable of capturing long-range dependencies and outperformed all other traditional ML models. We found that the TCN model has superior performance than other models with a model efficiency of 99.1\% in terms of NSE. In the model development phase input selection is an important task, but it is often neglected. In a study by \cite{thapa2020snowmelt}, it was found that gamma testing helps to determine the appropriate input combination, so in this study, we have compared each model on different input types and evaluated that the models with M2 input have better performance which is shown in table \ref{tab:model_comparison}. However, the TCN model with M1 input has a better model efficiency while all other models are performing better in M2 input which might be due to variations in spatial and temporal resolution or the relevance of the input variable.

Ground truth observation in the Himalayan basins is a difficult task due to the high elevation difference between the stations. Hence, the remotely sensed SCA and meteorological products are important assets.  This study proves the applicability of the ML model in snowmelt forecasting using remotely sensed SCA in data-scant basins. This study signifies that the SOTA models are more applicable for snowmelt forecasting as they capture long-range dependencies and are more generalizable.
\section{Conclusion}
In this work, we have investigated snowmelt-driven streamflow forecasting using ML techniques. The experiments were conducted over the HKH area. We compared the performance of SVR, LSTM, Transformer, and TCN architectures in M1 and M2 input for forecasting snowmelt. We have found that using M1 and M2 input does not give any significant performance differences also in this research work we have highlighted the importance of the hyperparameter tuning to obtain a better model. The TCN architecture showcased superior performance compared to other ML techniques. Furthermore, we disclosed the benefits of nested CV for performance evaluation on different ML techniques for snowmelt-driven streamflow forecasting. This contribution provides valuable insights for decision-makers in environmental monitoring and resource management, emphasizing the importance of leveraging advanced deep learning architecture for more reliable and precise forecasting.

\bibliographystyle{unsrtnat}
\bibliography{references}  






\end{document}